\documentclass[twocolumn]{article}

\usepackage{arxiv}
\usepackage[utf8]{inputenc} 
\usepackage[T1]{fontenc}    
\usepackage[citecolor=red]{hyperref}       
\usepackage{url}            
\usepackage{booktabs}       
\usepackage{amsfonts}       
\usepackage{nicefrac}       
\usepackage{microtype}      
\usepackage{cleveref}       
\usepackage{lipsum}         
\usepackage{graphicx}
\usepackage{doi}

\usepackage{multicol}
\title{Enhancing Self-Supervised Learning for Remote Sensing with Elevation Data: A Case Study with Scarce And High Level Semantic Labels}

\date{}

\author{ Omar A. Castaño-Idarraga \\
    Faculty of Engineering\\
	University of Antioquia\\
    \And
    Freddie Kalaitzis\\
    Department of Computer Science\\
    University of Oxford
	\And
	Raul Ramos-Pollán\\
    Faculty of Engineering\\
    University of Antioquia\\
}

\renewcommand{\headeright}{}
\renewcommand{\undertitle}{}
\renewcommand{\shorttitle}{}

\hypersetup{
pdftitle={A template for the arxiv style},
pdfsubject={q-bio.NC, q-bio.QM},
pdfkeywords={First keyword, Second keyword, More},
}

\begin{document}

\twocolumn[
  \begin{@twocolumnfalse}
    \maketitle
    \begin{abstract}
This work proposes a hybrid unsupervised and supervised learning method to pre-train models applied in Earth observation downstream tasks when only a handful of labels denoting very general semantic concepts are available. We combine a contrastive approach to pre-train models with a pixel-wise regression pre-text task to predict coarse elevation maps, which are commonly available worldwide. We hypothesize that this will allow the model to pre-learn useful representations, as there is generally some correlation between elevation maps and targets in many remote sensing tasks. We assess the performance of our approach on a binary semantic segmentation task and a binary image classification task, both derived from a dataset created for the northwest of Colombia. In both cases, we pre-train our models with 39k unlabeled images, fine-tune them on the downstream tasks with only 80 labeled images, and evaluate them with 2944 labeled images. Our experiments show that our methods, GLCNet+Elevation for segmentation, and SimCLR+Elevation for classification, outperform their counterparts without the pixel-wise regression pre-text task, namely SimCLR and GLCNet, in terms of macro-average F1 Score and Mean Intersection over Union (MIoU). Our study not only encourages the development of pre-training methods that leverage readily available geographical information, such as elevation data, to enhance the performance of self-supervised methods when applied to Earth observation tasks, but also promotes the use of datasets with high-level semantic labels, which are more likely to be updated frequently. Project code can be found in this link \href{https://github.com/omarcastano/Elevation-Aware-SSL}{https://github.com/omarcastano/Elevation-Aware-SSL}.
    \end{abstract}
  \end{@twocolumnfalse}

\vspace{10px}

]


\section{Introduction}

Inspired by the success of self-supervised learning methods in remote sensing \cite{vincenzi2021color, jean2019tile2vec, jung2021self, stojnic2021self, swope2021representation, jung2021contrastive}, we explore their potential for application to a dataset with and high-level semantic labels. Recent studies have shown that self-supervised learning methods can perform comparably or even better than their supervised counterparts in tasks such as image classification, object detection, and semantic segmentation on remote sensing data. Existing applications of self-supervised learning in remote sensing primarily utilize benchmark datasets with low-level semantic concepts. However, there is a significantly larger amount of high-level semantic labels available for remote sensing, which are more likely to be updated frequently. The application of self-supervised learning methods in this context appears promising, and a few works are emerging in this area \cite{la2022learning, bolyn2022mapping}, adapting learning from label proportions methods \cite{scott2020learning, tsai2020learning} to specific use cases. However, its application to remote sensing is still incipient.

In this direction, we first create a dataset comprising high-level semantic segmentation maps (refer to \ref{semantic_labels_predicctions}) for the northwest region of Colombia. This dataset, which we have named \href{https://huggingface.co/datasets/Ocastano/NWRC/tree/main}{Northwest Region Colombia Dataset} (NWRC dataset), is based on Sentinel-2 \cite{sentinel-2l} for the satellite images, and SIPRA \cite{sipra}, the Colombian agency for agricultural data and planning, for the semantic segmentation maps. We then use the NWRC dataset to evaluate the performance of two existing self-supervised contrastive learning methods, GLCNet \cite{glcnet} for a segmentation downstream task, which aims to predict the segmentation maps extracted from SIPRA, and SimCLR \cite{simclr} for a classification downstream task derived from the segmentation masks. This classification task can be seen as a spatially coarser counterpart of the segmentation task by setting the classification target to summarize the pixel classes. We find that, even with only high-level semantic labels available, these self-supervised methods outperform models trained from scratch when the amount of labeled data is limited.

Next, we propose a novel unsupervised learning approach that combines contrastive learning and elevation maps (Figure \ref{proposed-architecture}). Our approach involves designing a pixel-level regression pre-text task that predicts elevation maps at a coarser spatial resolution than the input images. This is then combined with SimCLR and GLCNet through an integrated loss function. We refer to these new frameworks as SimCLR+Elevation and GLCNet+Elevation, respectively. The intuition behind this approach is that there may be a correlation between elevation and the classes in the final downstream tasks. This correlation may enable the model to learn a useful representation that incorporates information about the classes in the final task.

In both cases, we pre-train our models with 39K unlabeled images, fine-tune the downstream task only with 80 labeled images, and test it with 2944 labeled images. Our experiments show that our proposed methods for pre-training, SimCLR+Elevation for classification tasks and GLCNet+Elevation for segmentation tasks, outperform both the SimCLR and GLCNet methods respectively, in terms of accuracy and macro average F1. This supports the idea that the inclusion of additional information, during the pre-training phase, which is associated with the classes in the downstream tasks, can boost model performance.

In summary, our contributions are: the creation and open-source release of a novel \href{https://huggingface.co/datasets/Ocastano/NWRC/tree/main}{dataset} for the northwest region of Colombia (the NWRC dataset), designed specifically for two tasks: classification and semantic segmentation. (2) the application of self-supervised contrastive learning to remote sensing using the NWRC dataset, demonstrating its ability to enhance model performance, even with only high-level semantic labels available, when the amount of labeled data is limited. (3) a novel unsupervised method that incorporates spatially coarse elevation maps as a pre-text task to boost the performance of current contrastive learning methods. 

\section{Background}
\label{sec:headings}

In this section, we present a brief overview of the contrastive framework for unsupervised learning and we describe two well-known self-supervised frameworks SimCRL\cite{simclr} and GLCNet\cite{glcnet}. It is important to highlight that in this research, SimCRL was consistently used to pre-train the models for the image classification downstream task. In contrast, GLCNet was used for pre-training models specifically designed for the image semantic segmentation downstream task.

\subsection{Contrastive Learning Framework}
\label{Contrastive Learning Framework}

Self-supervised contrastive learning is a type of unsupervised learning methodology that is used for representation learning. The objective in contrastive learning is to push the representations of positive samples closer together while pulling the representations of negative samples further apart. The positive samples are pairs of augmented versions of the same sample, while the negative samples are pairs of samples from different instances. Formally, given an unlabeled image $x$, two augmented views are generated, $\hat x$ and $x^+$. An encoder $e$ maps the input images into an embedding space, then a multilayer perceptron (MLP) $h$ is used to project the embeddings to a metric space, $\hat z = h(e(\hat x))$ and $z^+ = h(e(x^+))$. The encoder is trained to maximize the similarity between $\hat z$ and $z^+$, while minimizing the similarity between  $\hat z$ and $z^-$, where $z^-$ is randomly selected from a set of instances distinct from $x$. With the similarity measured by dot products, recent approaches in contrastive learning differ in the type of contrastive loss and generation of positive and negative pairs. In this work, we focus on SimCLR\cite{simclr} and GLCNet\cite{glcnet}, two well-known contrastive methods designed for image classification and image segmentation, respectively.

\begin{figure*}[h]
    \centering
    \includegraphics[width=450px]{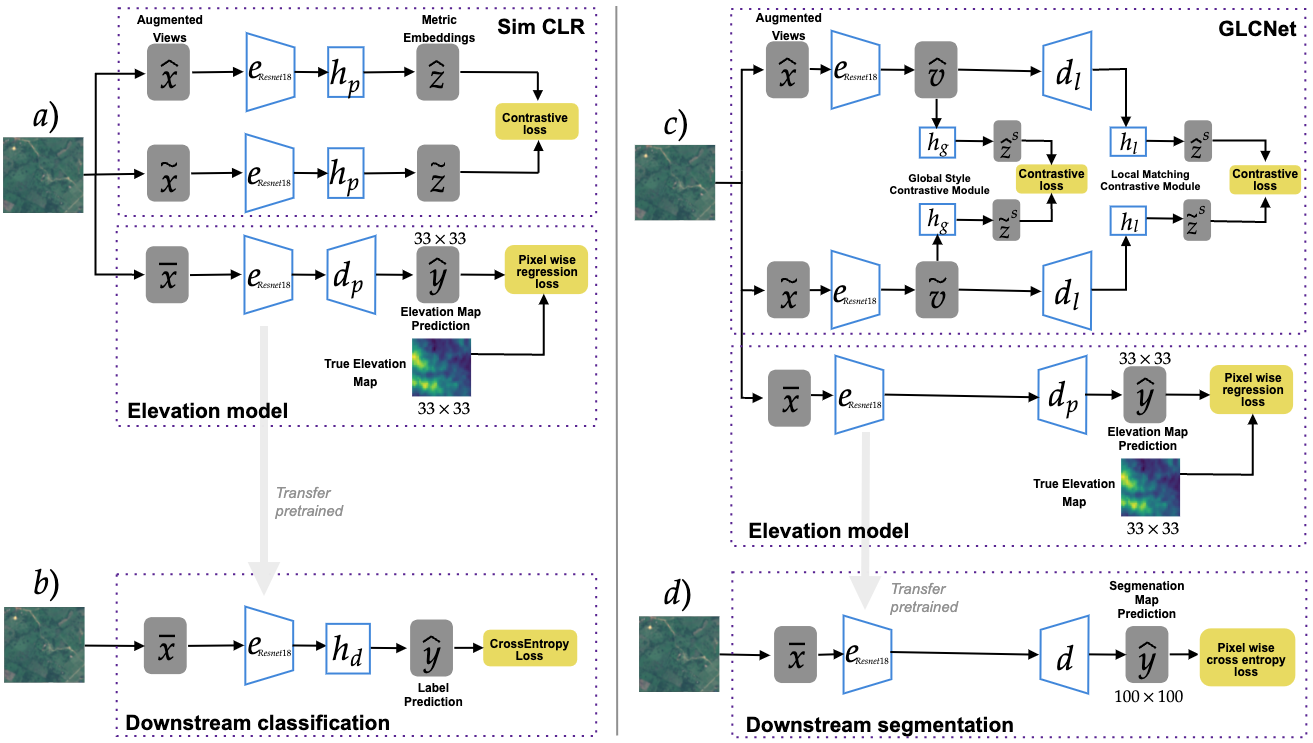}
    \caption{ a) Combined SimCLR+Elevation framework. b) Downstream classification using pre-trained backbone. c) Combined GLCNet+Elevation framework. d) Downstream segmentation using pre-trained backbone. Observe that in both (with SimCLR or with GLCNet) cases the Resnet18 encoder is shared during pre-training and it is only that encoder that is transferred to the downstream tasks. The rest of the architectural elements (the projection head $h_d$ and the decoder $d$) are initialized randomly.} 
    \label{proposed-architecture}
\end{figure*}

\subsubsection{SimCLR} 
SimCLR \cite{simclr} is a simple contrastive learning framework  where the representations are learned by maximizing agreement of positive samples, while minimizing agreement of negative samples. The positive samples are pairs of augmented versions of a single sample, while the negative samples are pairs of samples from different instances. In this framework, $N$ samples from a mini-batch are augmented to create $2N$ samples. A pair of samples derived from the same sample forms a positive pair, while the remaining $2(N-1)$ samples form negative samples. The contrastive loss, which is called \textit{NT-Xent}, is calculated as follows:
\begin{equation}\label{NTXent}
\mathcal{L}_{C}=\frac{1}{2 N} \sum_{k=1}^N\left(\ell\left(\tilde{x}_i, \hat{x}_i\right)+\ell\left(\hat{x}_i, \tilde{x}_i\right)\right)
\end{equation}
with: 
\begin{equation}
\resizebox{.11\vsize}{!}{$\ell\left(\tilde{x}_i, \hat{x_i}\right)=-\log$}\resizebox{.21\vsize}{!}{$\frac{\exp \left(sim\left(\tilde{z}_i, \hat{z}_i\right) / \tau\right)}{\sum_{x \in \Lambda^{-}} \exp \left(sim\left(\tilde{z}_i, h(e(x))\right) / \tau\right)}$}
\end{equation}

and 

\begin{equation}
    \hat z = h(e(\hat x)), \tilde{z} = h(e(\tilde{x}))
\end{equation}

where $e$ denotes the encoder used to map input images into an embedding space, $h$ denotes the MLP used to project embeddings into a metric space, $sim$ denotes the similarity measure function between two feature vectors, which in this case is cosine similarity. $\Lambda^{-}$ refers to the $2(N-1)$ negative samples in addition to the positive sample pair. $\tau$ is a temperature parameter. Figure \ref{proposed-architecture}-a illustrates the SimCLR framework.

\subsubsection{GLCNet} GLCNet\cite{glcnet} is a self-supervised learning method that aims to pre-train an encoder-decoder network for image semantic segmentation. It achieves this by combining two modules. The Global Style Contrastive Learning focuses on learning global features by contrasting different augmented views of the same image. It uses style features instead of the average pooling features commonly used in contrastive learning, which are believed to better represent the overall image characteristics. The Local Matching Contrastive Learning targets learning local features by contrasting small matching regions from different augmented views of the same image. It utilizes the output of the decoder to extract these local regions and employs a projection head to generate metric embeddings. The total loss function of GLCNet is based one the \textit{NT-Xent} loss function \ref{NTXent}, and combines the losses from both modules, with a weighting parameter $\lambda$ that controls their relative importance.

\begin{equation}\label{glcnetloss}
\mathcal{L}_{C}=\lambda\mathcal{L}_{G} + (1-\lambda) \mathcal{L}_{L}
\end{equation}

where $\mathcal{L}_{G}$ and $\mathcal{L}_{L}$ denote the global and local loss functions respectively. Both loss functions are the \textit{NT-Xent} loss function applied to the metric embeddings coming from the encoder and decoder, respectively. Figure \ref{proposed-architecture}-c illustrates the GLCNet framework.

\section{Method}
\label{sec:method}

This section introduces our unsupervised learning approach that leverage elevation maps for model pre-training. This approach is incorporated with contrastive learning frameworks, specifically SimCLR\cite{simclr} and GLCNet\cite{glcnet}. The objective is to improve their performance when the pre-trained backbone is applied to downstream tasks, particularly classification and semantic segmentation.

\subsection{Elevation Map Prediction as a Pre-text Task}
In this section, we explore the potential of using spatially coarse elevation maps for unsupervised learning. Specifically, we design a pixel-wise regression pre-text task that leverages elevation maps for pre-training a backbone. We input 100x100 pixel RGB images, and the model outputs 33x33 regression values for elevation. As this pre-text task is at the pixel level, we hypothesize that it may help the model to pre-learn local patterns that could be useful in a downstream segmentation task. Figure \ref{proposed-architecture}-a (Elevation model) illustrates our proposed pixel-wise regression pre-text task.

More specifically, given an image $x$, we use an encoder $e$ to map images into an embedding space of lower dimension, then we use a decoder $d$ to upscale the embedded representation and predict the elevation map, $\hat y = d(e(x))$, which might be of lower resolution than the input image $x$. This encoder-decoder network is based on the Unet \cite{unet} architecture. The encoder and decoder are optimized using the pixel-level root mean squared error loss function, which is calculated as follows:

\begin{equation}\label{pixel-level-regression-loss}
    \mathcal{L}_{E} = \frac{1}{N}\sum_{i=1}^N \sum_{j=1}^{W} \sum_{k=1}^{H} (y_{ijk} - d(e(x_i))_{jk} )^2
\end{equation}

Where $y_{jk}$ and $d(e(x_i))_{jk}$ are the ground truth and predicted values for the $i^{th}$ image at location $(j, k)$, respectively. N denotes the number of instances in the mini-bath, $W$ and $H$ denote the width and the high of the ground truth and predicted elevation map.  

Following the completion of the pre-training process, the decoder and projection header are eliminated, and only the encoder $e$ is utilized to initialize the backbone in the downstream task.

\subsection{Combining Elevation Prediction and Contrastive Learning}

In the previous subsection, we designed a regression pre-text task leveraging elevation maps. In this section, we aim to integrate the elevation map prediction pre-text task with the contrastive learning frameworks, SimCLR and GLCNet, in order to improve the performance of these models.

\subsubsection{SimCLR+Elevation}
This integrated framework combines the pixel-wise regression pre-text task, presented in the previous subsection, with the contrastive task proposed in SimCLR (refer to Figure \ref{proposed-architecture}, upper-left).The framework involves generating three augmented views of an image, namely $\hat{x}$, $\tilde{x}$, and $\overline{x}$. These augmented views are then mapped to a shared embedding space using an encoder $e$, resulting in $\hat{v}=e(\hat{x})$, $\tilde{v}=e(\tilde{x})$, and $\overline{v}=(\overline{x})$. For the contrastive pre-text task, the embedded views $\hat{v}$ and $\tilde{v}$ are passed through a projection header $h_p$, which is a multi-layer perceptron, resulting in metric space representations $\hat{z}=h_p(\hat{v})$ and $\tilde{z}=h_p(\tilde{z})$ which are used in the contrastive pre-text task to learn global features. On the other hand, the remaining embedded view $\overline{v}=e(\overline{x})$ is utilized to predict the elevation map. To accomplish this, a decoder $d$ is employed to upscale the embedded representation and predict the elevation map $\hat y = d_p(e(x))$. The network is jointly optimized to perform both contrastive learning and elevation map prediction. Therefore, a loss function combining the NT-Xent loss (defined in Equation \ref{NTXent}) and the pixel level MSE (define in Equatio \ref{pixel-level-regression-loss}) loss with a coefficient $\alpha$ is introduced. The loss function is defined as:

\begin{equation}\label{proposedloss}
\mathcal{L} = \alpha\mathcal{L}_{E} + (1 - \alpha)\mathcal{L}_{C}
\end{equation}

Here, the coefficient $\alpha$ represents the relative importance of the contrastive learning loss $\mathcal{L}_{C}$ and the elevation map learning loss $\mathcal{L}_{E}$. The joint optimization of both tasks allows our model to learn representations that simultaneously maximize agreement between positive image pairs, minimize agreement between negative image pairs, and accurately predict the elevation map of the images.

Upon the completion of the pre-training process using the combined framework, SimCLR+Elevation, the decoder $d_p$ and projection header $h_p$ are discarded, and only the encoder $e$ is used to initialize the backbone in the classification downstream task.

\subsubsection{GLCNet+Elevation}
Similar to SimCLR+Elevation, the GLCNet+Elevation framework merges the regression pre-text task with the contrastive learning mechanism of GLCNet (illustrated in Figure \ref{proposed-architecture}, upper-right). This approach also generates three augmented images: $\hat{x}$, $\tilde{x}$, and $\overline{x}$, which are encoded into a shared embedding space. The GLCNet framework introduces two contrastive modules: the global style contrastive module and the local matching contrastive module. For the global style module, the embeddings $\hat{v}$ and $\tilde{v}$ are processed by a projection head $h_g$, resulting in metric space representations $\hat{z}=h_g(\hat{v})$ and $\tilde{z}=h_g(\tilde{v})$, which are used to learn global features. For the local matching module, the embeddings $\hat{v}$ and $\tilde{v}$ are decoded and then projected to obtain metric space embeddings $\hat{z}=h_l(d_l(\hat{v}))$ and $\tilde{z}=h_l(d_l(\tilde{v}))$, which are used to learn local features beneficial for segmentation tasks. The elevation map prediction uses the embedded view $\overline{v}$, with a decoder $d_p$ predicting the elevation map $\hat y = d_p(e(\overline{x}))$. The network is jointly optimized for contrastive learning and elevation map prediction, with a loss function that combines the loss proposed in GLCNet \ref{glcnetloss} and the pixel-level MSE loss \ref{pixel-level-regression-loss}, weighted by coefficients $\alpha$ and $\lambda$, as shown in Equation \ref{proposedloss}.

\begin{equation}\label{proposedloss-glcnet}
\mathcal{L} = \alpha\mathcal{L}{E} + (1 - \alpha)(\lambda\mathcal{L}{G} + (1-\lambda) \mathcal{L}_{L})
\end{equation}

Here, the coefficient $\alpha$ serves to balance the elevation prediction loss $\mathcal{L}_{E}$ with the global and local contrastive losses, $\mathcal{L}_{G}$ and $\mathcal{L}_{L}$, respectively. The parameter $\lambda$ further adjusts the emphasis between the global and local contrastive learning objectives.

Upon the completion of the pre-training, the network is fine-tuned on the downstream task. Specifically, the decoders $d_l$ and $d_p$, and the projection heads $h_g$ and $h_l$ are removed, and the pre-trained encoder $e$ is used to initialize the backbone for the downstream segmentation semantic task.

\section{Data Description}\label{data-description}

In this study, we developed a dataset for the northwest region of Colombia by utilizing Sentinel-2\cite{sentinel-2l} imagery and the agricultural land frontier defined by SIPRA\cite{sipra}, the official spatial viewer for the agricultural sector in Colombia. The dataset includes 42704 images of 100x100 pixels at 10m per pixel, featuring the RGB spectral bands. The labels are segmentation maps with two classes representing the general concept of:

\textbf{National agricultural frontier} (\emph{farmland}): areas where agricultural activities such as agriculture, livestock, aquaculture and fishing take place. This general 
label includes various fine-grained labels such as vegetable crops, tuber crops, confined crops, lagoons, lakes, natural swamps, wooded pastures, and mosaic of pastures and crops among others, which we do not have access to.

\textbf{Natural forest and non-agricultural areas} (\emph{other} or \emph{not farmland}): areas where agricultural activities do not take place. This general label includes various fine-grained labels such as buildings, roads, trails, plains, and natural forests among others, which we do not have access to.

We also use the geo-reference information of the images collected from Sentinel-2 to extract 30m elevation maps of 33x33 pixels from SRTMGL1Nv003\cite{nasa-maps}. These elevation maps will be used in our proposed pixel-wise regression pre-text task. 

\subsection{Semantic Segmentation Dataset}
\label{Semantic Segmentation Dataset}
The semantic segmentation dataset is formed using the segmentation maps which provide pixel-level annotations for the two principal categories: \textit{National agricultural frontier} and \textit{Natural forest and non-agricultural areas} (refer to Figure \ref{semantic_labels_predicctions} on the Ground Truth column). To benchmark the performance of unsupervised models (GLCNet and GLCNet+Elevation) with a limited amount of labeled data for fine-tuning, we randomly selected 3024 images. Of these, 80 images, representing a mere 0.2\% of the pre-training dataset, are designated for fine-tuning, while the remaining 2944 images are used for testing. The remaining 39720 images are left for unsupervised pre-training.

\subsection{Classification Dataset}
\label{Classification Dataset}
\begin{figure}[ht]
\centering
\includegraphics[width=200px, height=100px]{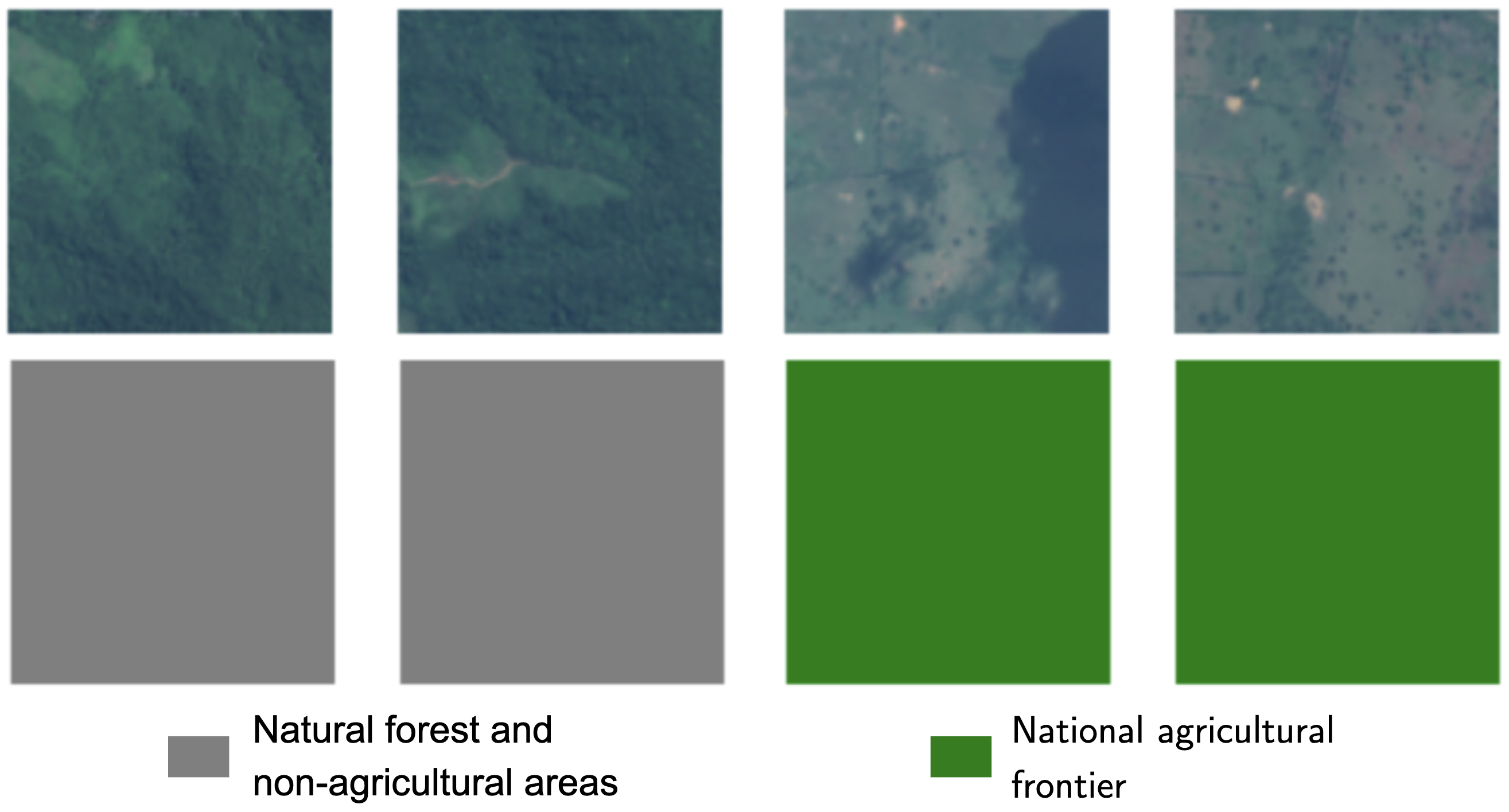}
\caption{Samples from the NWRC dataset used in the image classification task. The classification task involves predicting a label rather than a 2D segmentation map.}
\label{fig:clf-dataset}
\end{figure}

The classification dataset is derived from the segmentation maps containing pixel-level class information. For this dataset, we chose images that exclusively depict one class, as shown in Figure \ref{fig:clf-dataset}. Unlike the task of predicting a 2D segmentation map, the goal here is to assign a single label to each image, either \textit{National agricultural frontier} or \textit{Natural forest and non-agricultural areas}, following the standard practice in classification tasks. To evaluate the efficacy of unsupervised learning techniques (SimCLR and SimCLR+Elevation) with a limited labeled dataset for fine-tuning, we randomly selected 3024 images. From this subset, 80 images, which constitute 0.2\% of the pre-training dataset, are used for fine-tuning, and the remaining 2944 images are set aside for testing. The bulk of the dataset, which includes $39720$ images, is reserved for unsupervised pre-training.

\section{Experiments and Results}\label{sec:Experiments}

We evaluated the performance of our unsupervised pre-training approach on the image classification and semantic segmentation datasets described earlier. For image classification, we pre-trained a backbone using SimCLR+Elevation and compared its performance against three baselines. The first baseline, referred to as Random-Init, denotes a model that was trained from scratch on the classification dataset without any pre-training. The second, termed Elevation-Map, represents a model pre-trained solely on pixel-wise regression pre-text tasks. The third baseline, known as Vanilla SimCLR, is a model pre-trained using the standard SimCLR framework.

For semantic segmentation, we evaluated the performance of our unsupervised GLCNet+Elevation model in a similar manner. The Random-Init baseline in this context refers to a model trained from scratch on the segmentation dataset, without any pre-training. The Elevation-Map comparison involves a model that has been pre-trained solely on the elevation regression task. Finally, the Vanilla GLCNet baseline denotes a model pre-trained using the GLCNet approach.

\subsection{Image Classification on the NWRC Dataset}\label{classification}
We conducted experiments on the NWRC classification dataset presented in subsection \ref{Classification Dataset}. Initially, we pre-trained a backbone using an unsupervised strategy, including SimCLR, Elevation-Map, and SimCLR+Elevation, on 39,720 unlabeled images. This pre-training phase was followed by a fine-tuning process on a significantly smaller dataset of 80 labeled images. The performance of these fine-tuned models was then assessed using a test set of 2,944 labeled images.

\subsubsection{Implementation Details for Unsupervised Learning}\label{unsupervised-learning-clf}
For image classification, we employed SimCLR\cite{simclr} as the backbone of our contrastive learning method. We generated positive and negative pairs for network pre-training using various image augmentation techniques, such as color jittering, random grayscale conversion, horizontal and vertical flipping, and random resized cropping. We chose the ResNet18 architecture for the backbone and pre-trained the network for 200 epochs with a batch size of 256. The Adam optimizer was used with a weight decay of 1e-4, an initial learning rate of 0.001, and a cosine decay learning rate schedule. A temperature scaling $\tau$ of 0.5 was applied to calculate the contrastive loss. For the pixel-wise regression pre-text task, we also applied image augmentations, including color jittering, random grayscale conversion, and flipping, and a Unet architecture was used to predict the elevation maps. When integrating the elevation map task with contrastive learning, we set the weight $\alpha$ to 0.5 in the combined loss function.

\subsubsection{Implementation Details for Fine-Tuning}\label{Implementation Details for Fine-Tuning}
We evaluated the pre-learned representations by training a linear classification layer on top of the pre-trained backbone using 80 labeled data points. More specifically, we initialized a ResNet18 backbone with a pre-trained representation (SimCLR, Elevation-Map, and SimCLR+Elevation) and added a single fully-connected layer that maps the intermediate representation (embeddings) to class logits. We then proceeded to freeze the pre-trained backbone and train only the added layer for 20 epochs using a batch size of 8 and the Adam optimizer with an initial learning rate of 1e-3. Finally, we fine-tuned the entire network for 80 epochs with a batch size of 8 and a learning rate of 1e-5. Figure \ref{proposed-architecture} (upper-left) illustrates the configuration of our proposed approach.

\subsubsection{Quantitative Analysis}
The results of our classification experiments on the NWRC dataset are presented in Table \ref{tab:CLF-SIPRA}. We can see that the SimCLR+Elevation initialization method outperforms other initialization methods in terms of accuracy and macro average F1. In particular, the SimCLR+Elevation initialization outperforms Random initialization (Random-Init) and SimCLR by $6.2\%$ and $2.7\%$ F1, respectively. This confirms our hypothesis that incorporating features correlated with the classes in the pre-training phase can boost the performance of contrastive methods, in this case, SimCLR. However, we also notice that the Elevation initialization (Elevation-Map), which corresponds to setting $\alpha=1.0$ in the SimCLR+Elevation framework (Equation \ref{proposedloss}), does not perform better than Random-Init. This may be due to the fact that the pixel-wise regression pre-text task is not well-aligned with the downstream classification task and there is not enough data to fine-tune the network effectively for classification.

\begin{table}[h]
\centering
\begin{tabular}{c|c|c}
\hline

\textbf{Pre-train} & \textbf{Accuracy} & \textbf{F1} \\ \hline \hline
Random-Init        & 87.73             & 86.85             \\ \hline
SimCLR             & 90.79             & 90.31             \\ \hline
Elevation-Map (ours)          & 87.64             & 86.46             \\ \hline
SimCLR+Elevation (ours)   & \textbf{93.5}     & \textbf{93.04}    \\ \hline
\end{tabular}
\caption{\label{tab:CLF-SIPRA}Comparison of fine-tuning accuracy and F1-score on the classification test set for different pre-training methods. The ResNet18 backbone was pre-trained on 39,720 images. The SimCLR+Elevation method (ours) outperforms the other methods.}
\end{table}

\begin{figure*}[h]
    \centering
    \includegraphics[width=370px, height=310px]{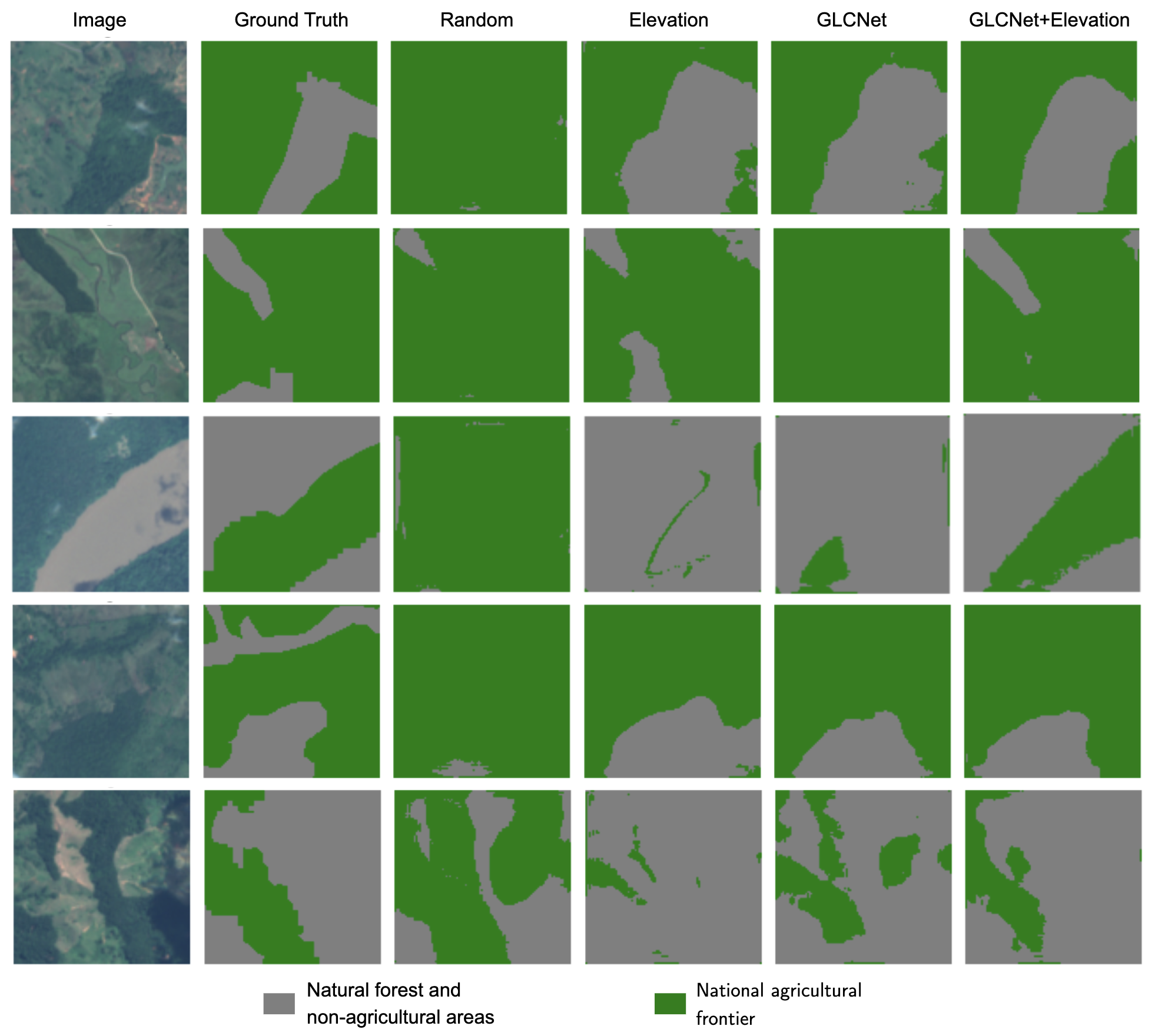}
    \caption{This Figure shows the experimental results of all models and the ground truth on the semantic segmentation dataset.} 
    \label{semantic_labels_predicctions}
\end{figure*}

\subsection{Semantic Segmentation on the NWRC Dataset}

We further evaluated our pre-training approach on the semantic segmentation task using the dataset described in subsection \ref{Semantic Segmentation Dataset}. A backbone was pre-trained using unsupervised methods, including GLCNet, Elevation-Map, and GLCNet+Elevation, on 39,720 unlabeled images, followed by fine-tuning on 80 labeled images. The performance of the fine-tuned models was then evaluated on a test set of 2,944 labeled images, using the F1 score and mean intersection over union (MIoU) as metrics.

\subsubsection{Implementation Details for Unsupervised Learning}\label{unsupervised-learning-segmentation}
In our semantic segmentation experiments, we used GLCNet as the backbone for contrastive learning. To pre-train the network using contrastive learning, we generated positive and negative pairs using various image augmentation techniques, including color jittering, random grayscale, horizontal and vertical flipping, and random resized cropping. We pre-trained the network for 200 epochs with a batch size of 200 using a ResNet18 architecture, employing the Adam optimizer with a weight decay of 1e-4, an initial learning rate of 0.001, and a cosine decay schedule. For the elevation map regression pre-text task, we applied image augmentations and used a Unet architecture to predict the elevation maps. We set $\alpha=0.5$ (Equation \ref{proposedloss-glcnet}) when integrating the pixel-level regression task into contrastive learning.

\subsubsection{Implementation Details for Supervised Learning}
To test the pre-trained representations, we used a Unet architecture with the pre-trained ResNet18 backbones. We initialized and froze the Unet's encoder with the pre-trained ResNet18 and trained the decoder for 20 epochs with a batch size of 8, using the Adam optimizer at a learning rate of 1e-3. The entire network was then fine-tuned for 80 epochs with a batch size of 8 and a learning rate of 1e-5. Additional data augmentation techniques were applied to the training images. Figure \ref{proposed-architecture} (upper-right) illustrates our approach for pre-training and fine-tuning on a semantic segmentation task.

\subsubsection{Quantitative Analysis}
The semantic segmentation results are summarized in Table \ref{tab:SEG-SIPRA}. The GLCNet+Elevation method achieved the highest F1 score and MIoU, outperforming Random and GLCNet by $4.29\%$ and $1.54\%$ in MIoU, respectively. This supports the notion that pre-training with class-correlated features can enhance model performance in downstream tasks. Additionally, we observe that the Elevation initialization, which corresponds to setting $\alpha=1.0$ in the GLCNet+Elevation framework, performed better than Random initialization and was close to GLCNet, suggesting that the alignment between the regression pre-text task and the semantic segmentation task is beneficial. Visual comparisons in Figure \ref{semantic_labels_predicctions} further illustrate the superior performance of our GLCNet+Elevation method compared to other approaches.

\begin{table}[h]
\centering
\begin{tabular}{c|c|c}
\hline

\textbf{Pre-train} & \textbf{F1} & \textbf{MIoU} \\ \hline \hline
Random Init        & 71.81            & 56.72            \\ \hline
GLCNet             & 74.17             & 59.47
             \\ \hline
Elevation-Map (ours)          & 73.29            & 58.0
             \\ \hline
GLCNet+Elevation (ours)   & \textbf{75.33}     & \textbf{61.01}    \\ \hline
\end{tabular}
\caption{\label{tab:SEG-SIPRA} Fine-tuning MIoU and F1-score on the semantic segmentation test set. We use a ResNet18 backbone pre-trained on 39720 images.}
\end{table}

\subsection{Ablation Analysis}

We conducted a series of ablation experiments to investigate the influence of increasing the quantity of labeled data for the fine-tuning of pre-trained models in both image classification and semantic segmentation tasks. Specifically, we adjusted the volume of labeled data utilized for fine-tuning the pre-trained models on the downstream tasks within the range of $[80, 100, 200, 500, 1000, 2000, 3000]$, while using 36,800 images for pre-training and 2,944 images for testing.

\begin{figure}[h]
\centering
\includegraphics[width=225px, height=130px]{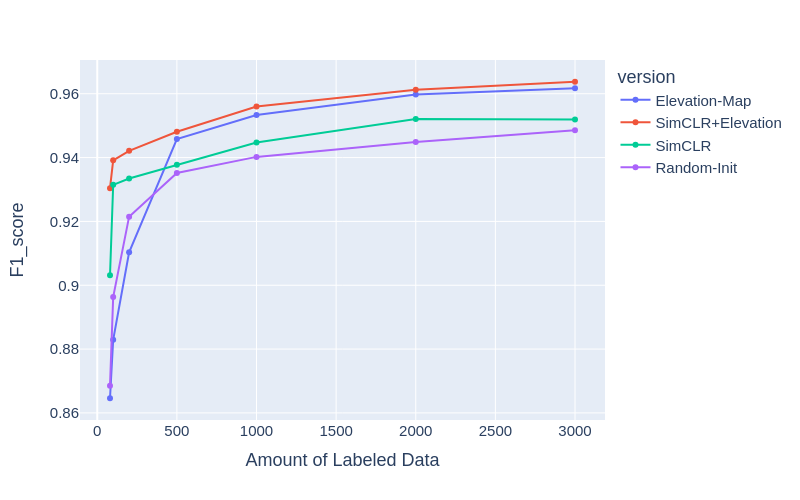}
\caption{Results from the ablation analysis showcasing the impact of varying amounts of labeled data on the fine-tuning of pre-trained models for image classification.}
\label{ablation-classification}
\end{figure}

In the context of image classification, our proposed methodology, SimCLR+Elevation, demonstrated superior performance over other unsupervised pre-training techniques across all tested scenarios, as depicted in Figure \ref{ablation-classification}. A more pronounced impact was observed when the amount of labeled data was less than 500. Additionally, SimCLR consistently outperformed Random-Init across all tested scenarios.

Interestingly, the Elevation-Map pre-training method underperformed in comparison to Random-Init and SimCLR when the amount of labeled data was less than 500, but surpassed them when the amount of labeled data was 500 or greater. We hypothesize that this is due to the Elevation pre-training being a regression pixel-level pre-text task, which does not align well with the final classification downstream task. Therefore, it requires more labeled data, compared to SimCLR or SimCLR+Elevation, to effectively leverage the pre-training.

\begin{figure}[h]
\centering
\includegraphics[width=225px, height=130px]{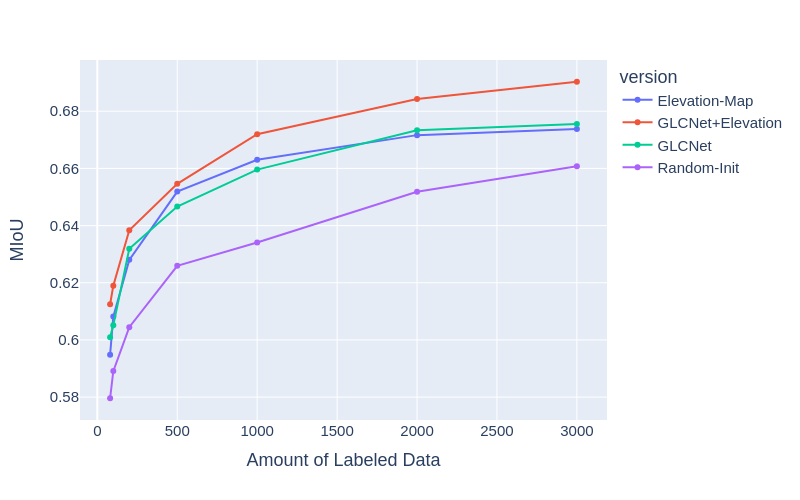}
\caption{Results from the ablation analysis illustrating the effect of different amounts of labeled data on the fine-tuning of pre-trained models for semantic segmentation.}
\label{ablation-segmentation}
\end{figure}

In terms of semantic segmentation, our proposed methodology, GLCNet+Elevation, outperformed other unsupervised pre-training techniques in all tested scenarios, as depicted in Figure \ref{ablation-segmentation}. Moreover, GLCNet and Elevation-Map pre-training methods demonstrated comparable performance and consistently surpassed Random-Init in all tested scenarios. In this instance, we hypothesize that as the regression pixel-level pre-text task aligns well with the final semantic segmentation downstream task, which is also a pixel-level task.

\section{Conclusion}
\label{sec:Conclusion}
In this study, we proposed a novel hybrid unsupervised and supervised learning method for pre-training models applied to Earth observation tasks. We combined a contrastive approach with a pixel-wise regression pre-text task to predict coarse elevation maps, hypothesizing that this would allow the model to pre-learn useful representations. Our experiments on a binary segmentation downstream task and a binary image classification task, both derived from a dataset for the northwest of Colombia, demonstrated that our methods, GLCNet+Elevation for segmentation, and SimCLR+Elevation for classification, outperformed their counterparts without the pixel-wise regression pre-text task, namely SimCLR and GLCNet, in terms of macro-average F1 Score and Mean Intersection over Union (MIoU).

Our findings suggest that leveraging readily available geographical information, such as elevation data, can enhance the performance of self-supervised methods when applied to Earth observation tasks. Furthermore, our study promotes the use of datasets with high-level semantic labels, which are more likely to be updated frequently.

Our ablation analysis also revealed that the amount of labeled data used for fine-tuning significantly influences the performance of pre-trained models. In particular, our proposed methods, SimCLR+Elevation and GLCNet+Elevation, consistently outperformed other methods across all tested scenarios, with a more pronounced impact observed when the amount of labeled data was less than 500.

In conclusion, our study contributes to the development of pre-training methods for Earth observation tasks, providing a promising approach that combines contrastive learning and elevation maps. Future work may explore the application of our proposed method to other geographical datasets and tasks, as well as the integration of other types of geographical information into the pre-training process.

\end{document}